\documentclass{article}
\usepackage{arxiv}
\usepackage[utf8]{inputenc}
\usepackage[T1]{fontenc}
\usepackage{graphicx}
\graphicspath{{./figures/}{./}}
\usepackage{amsmath,amssymb,amsfonts,bm,mathrsfs}
\usepackage{amsthm}
\usepackage{booktabs}
\usepackage{array,tabularx,multirow,threeparttable}
\usepackage{siunitx}
\usepackage{textcomp}
\usepackage{xcolor}
\usepackage{pifont}
\usepackage{url}
\usepackage{cite}
\usepackage{subcaption}
\usepackage{microtype}
\usepackage{hyperref}
\hypersetup{colorlinks=true,linkcolor=blue,citecolor=blue,urlcolor=blue}
\newcolumntype{C}{>{\centering\arraybackslash}X}

\title{A Differentiable Chassis Safety Layer for Intelligent Vehicle Motion Control}
\author{%
Lu Xiong$^{1,2}$ \quad Xuge Hu$^{2}$ \quad Yufei Song$^{2}$\\[0.5em]
Weidong Liu$^{1}$ \quad Chen Tang$^{2,*}$\\[0.9em]
{\small $^{1}$Key Laboratory of Road and Traffic Engineering of the Ministry of Education,}\\
{\small Tongji University, Shanghai, China}\\
{\small $^{2}$College of Automotive and Energy Engineering, Tongji University, Shanghai, China}\\[0.3em]
{\small $^{*}$Corresponding author: \href{mailto:chen_tang@tongji.edu.cn}{chen\_tang@tongji.edu.cn}}
}
\date{}
\begin{document}
\maketitle

\begin{abstract}
Learning-based vehicle motion control requires coordination between diverse task objectives and state-dependent chassis safety constraints. Conventional cascaded architectures may compromise task performance when safety filters modify nominal commands independently of policy learning. This article proposes a differentiable chassis safety layer that modularly integrates task-oriented control with explicit physical constraints. Actuator limits, individual-tire force constraints, and vehicle handling envelopes are incorporated into sequential quadratic programs. Each projected command propagates the predicted vehicle dynamics and updates subsequent constraints, enabling temporal gradient coupling through implicit differentiation. A two-stage learning strategy combines expert imitation with risk-guided supervision and conditional intervention guidance to coordinate nominal control decisions with chassis safety constraints. Closed-loop evaluations under unseen scenarios, emergency maneuvers, and ablation studies demonstrate that the proposed method improves safety–performance coordination and exploits vehicle dynamic potential while maintaining handling stability.
\end{abstract}

\keywords{
Vehicle Handling and Stability Control, Differentiable Quadratic Programming, Control Barrier Function, Differentiable Safety Projection, Joint Policy--Safety Optimization
}

\section{Introduction}
Intelligent vehicle motion control involves multiple objectives, including trajectory tracking, energy efficiency, ride comfort, and actuator performance \cite{iotj_zhao2026_path,11701498,11175236}. Achieving these objectives requires coordination with physical constraints imposed by vehicle dynamics, tire--road interaction, and actuator capabilities. In multi-actuator chassis systems, these constraints are coupled through vehicle motion and vary with operating conditions \cite{Bobier-Tiu02092019,9837329}. Therefore, integrating task optimization with chassis physical constraints remains a fundamental challenge.

\begin{figure}[!t]
    \centering
    \includegraphics[width=0.94\linewidth]{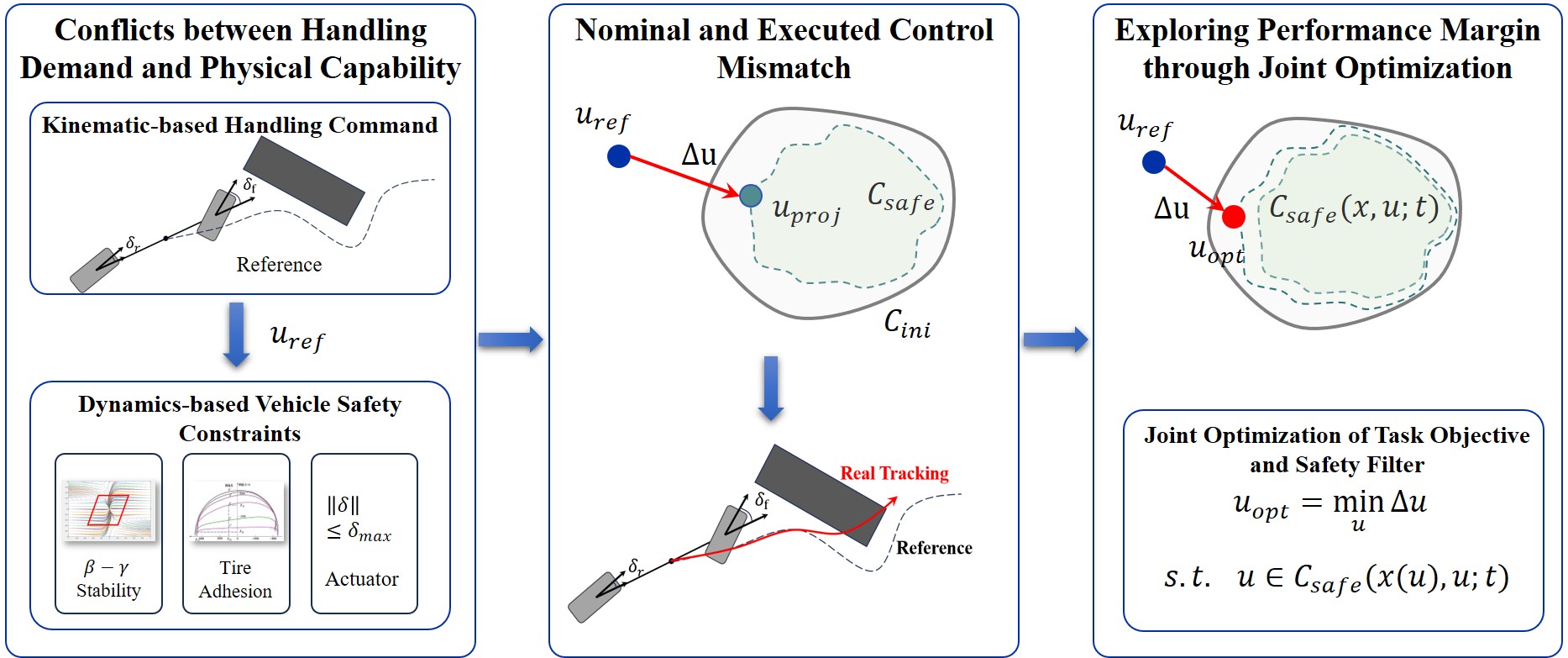}
    \caption{Motivation for policy--safety coordination. (a) Handling demand versus physical limits. (b) Nominal-to-executed steering mismatch after safety projection. (c) Coordination of steering policy learning and state-dependent safety correction. The admissible sets are schematic.
    }
    \label{fig:introduction}
\end{figure}
Model predictive control (MPC) explicitly incorporates physical constraints into task optimization, while learning-based extensions improve adaptation through data-driven models and parameters \cite{2020Learning,NingyuanGuo2022rea1}. Nevertheless, their effectiveness depends on prediction accuracy, particularly under nonlinear tire behavior, uncertain adhesion, and model mismatch \cite{9638389,lorenzen2017adaptive}. Alternatively, control barrier functions (CBFs) enforce state-dependent constraints through constrained optimization \cite{7782377}. High-order and predictive formulations extend this approach to more complex system dynamics \cite{9516971,zeng2021safety,10644741}. However, independently applied constraint projections may modify task-oriented commands without considering the resulting performance degradation \cite{hsu2023safety}.

Learning-based controllers provide flexibility in optimizing complex motion tasks \cite{9772321,10005026}. Recent applications include autonomous navigation, networked suspension control, and risk-aware decision-making \cite{iotj_yao2025_nsgvf,iotj_zhu2025_suspension,iotj_zhang2026_eirm}. To incorporate explicit constraints into neural learning, OptNet introduced differentiable optimization layers \cite{amos2021optnetdifferentiableoptimizationlayer,agrawal2019differentiable}. Adaptive CBF methods further enabled parameterized safety conditions \cite{9410332,9129764}, while BarrierNet integrated differentiable CBF-constrained QPs into neural controllers \cite{xiao2023barriernet}. However, optimization-layer gradients may become uninformative under certain active-constraint configurations \cite{pmlr-v235-paulus24a}.

For chassis motion control, the central difficulty lies in the dynamic coupling of heterogeneous physical constraints. Unlike geometric motion constraints, tire-force capacity depends on longitudinal and lateral force demands, vertical loads, and road adhesion, while handling limits depend on the evolving sideslip and yaw states. These state--control dependencies are particularly important during high-dynamic maneuvers. Moreover, each constrained control command changes subsequent vehicle states and constraint conditions. Consequently, independent single-step projections cannot explicitly represent the temporal interaction between successive chassis constraints within the learning process. This motivates a dynamics-coupled differentiable optimization architecture tailored to chassis motion control.

To address this challenge, this article proposes \textit{ChassisOptNet}, a differentiable chassis optimization network for joint task--constraint learning. The network integrates actuator limits, tire-force constraints, and vehicle handling envelopes into state-dependent quadratic programming (QP) layers. Successive QPs are coupled through predictive vehicle dynamics, enabling constraint corrections and their temporal sensitivities to influence task-oriented learning. A two-stage training strategy combines expert imitation with risk-guided supervision and conditional intervention guidance. The framework is demonstrated on four-wheel-steering (4WS) vehicle motion control, using trajectory tracking as a representative task.

The main contributions are summarized as follows:
\begin{itemize}
    \item \textbf{Chassis-specific differentiable safety layer:}
    A modular safety layer integrates actuator limits, individual-tire friction constraints, and CBF-based handling envelopes into a unified control-space QP formulation, accounting for their coupling with vehicle dynamics.

    \item \textbf{Dynamics-unrolled safety projection:}
    Sequential differentiable QPs are coupled through predictive vehicle dynamics, allowing each projected control action to update subsequent states and safety constraints. Implicit differentiation provides temporal projection sensitivities for upstream learning.

    \item \textbf{Safety-guarded task learning:}
    A two-stage learning strategy combines expert imitation with expert-relative risk supervision and conditional safety-intervention guidance, coordinating task-oriented control with chassis safety projection. Closed-loop simulations validate the framework under unseen and near-limit operating conditions.
\end{itemize}

% \begin{figure}[ht!]
% 	\centering
% 	\includegraphics[width=0.5\textwidth]{ITSC2014 4WS Parking/Figure/4wsCharacteristic.png}
% 	\caption{CrabWalk Maneuver for Enhanced Parking Efficiency}
%    \label{fig:4wsCharacteristic}
% \end{figure}
The remainder of this article is organized as follows. Section~\ref{sec:idcbf} develops the 4WS vehicle dynamics, handling-stability constraints, and dynamics-unrolled differentiable safety projection. Section~\ref{sec:Safety-Guaranteed Bi-Level Policy Optimization Framework} introduces the joint policy-learning framework and two-stage training procedure. Section~\ref{sec:Training and Online Testing Experiments} reports the closed-loop simulation results and ablation studies. Section~\ref{sec:discussion} discusses the findings and limitations. Section~\ref{sec:conclusion} concludes this article.

\section{Differentiable Chassis Safety Layer} 
\label{sec:idcbf}
Vehicle handling-stability constraints act on steering inputs, tire forces, and lateral--yaw states, which are coupled through front- and rear-wheel steering. This section maps these constraints into a unified control-space formulation and incorporates them into sequential differentiable safety projections. Each projected steering command updates the predicted vehicle state and subsequent constraints, enabling temporal gradient propagation through the safety projections and vehicle dynamics during policy training.

\subsection{Vehicle Dynamics for Safety Projection}
A two-degree-of-freedom bicycle model with linear tire cornering stiffness is employed to predict lateral and yaw motion. Longitudinal speed is regulated by a separate speed-tracking controller and is treated as an available parameter in the lateral controller. The state and control vectors are defined as $x=[v_y,\gamma]^T$ and $u=[\delta_f,\delta_r]^T$, where $v_y$ is lateral velocity, $\gamma$ is yaw rate, and $\delta_f$ and $\delta_r$ are the front- and rear-wheel steering angles. The control-affine model is given by
\begin{align}
    \dot{x} &= \mathcal{F}(x)+\mathcal{G}u \nonumber \\
    \mathcal{F}(x)&=
    \begin{bmatrix}
       -\dfrac{2(C_f+C_r)}{mv_x}v_y
       -\left(v_x+\dfrac{2(l_fC_f-l_rC_r)}{mv_x}\right)\gamma \\
       -\dfrac{2(l_fC_f-l_rC_r)}{I_zv_x}v_y
       -\dfrac{2(l_f^2C_f+l_r^2C_r)}{I_zv_x}\gamma
    \end{bmatrix}\nonumber\\
    \mathcal{G}&=
    \begin{bmatrix}
       \dfrac{2C_f}{m} & \dfrac{2C_r}{m}\\
       \dfrac{2l_fC_f}{I_z} & -\dfrac{2l_rC_r}{I_z}
    \end{bmatrix}.
    \label{eq:vehicle model}
\end{align}
Here, $m$ and $I_z$ denote vehicle mass and yaw moment of inertia, respectively. The distances from the center of mass to the front and rear axles are $l_f$ and $l_r$. The quantities $C_f$ and $C_r$ denote the cornering stiffness of an individual front and rear tire, respectively, and $v_x$ is the longitudinal velocity. The factors of two in \eqref{eq:vehicle model} account for the two tires on each axle. The model is used within the operating region $v_x>0$, with a positive numerical lower bound applied to longitudinal speed during implementation. Its linear tire representation provides a computationally tractable prediction model, while nonlinear tire saturation is addressed through additional friction constraints.

\subsection{Multi-Level Chassis Constraints}\label{sec:safety constraints}
Vehicle handling-stability constraints are classified into three categories according to their state–control dependence: control-space, state–control coupled, and state-space constraints. Actuator limits are directly imposed on steering commands, while tire-friction constraints are mapped into state-dependent steering limits through the tire model. Handling-envelope constraints are converted into control-dependent inequalities using control barrier functions (CBFs). These constraints form a unified control-space representation for the subsequent differentiable safety projection.
% 可微分安全
%Stability is a key aspect of ensuring vehicle safety\cite{Bobier-Tiu02092019}, where the tire adhesion ellipse \cite{9837329} and phase-portrait-based yaw stability region\cite{10147032} are essential constraints. This study constructs tire adhesion ellipse constraints and CBF-based yaw stability constraints.

% 为了确保车辆运行时的横向以及横摆稳定性，通常需要考虑轮胎附着椭圆约束以及横摆稳定域约束。轮胎附着椭圆约束要求车辆纵向力和侧向力的合力不能超过路面附着极限，其从单个车轮的角度确保轮胎不会出现严重滑移现象。横摆稳定性约束从整车的角度确保车身整体的运动状态处于可控的稳定极限内，本研究采用菱形稳定域构建横摆稳定性约束。
\subsubsection{Constraints in Control Space}
Steering-angle and steering-rate limitations act directly on the control vector $u=[\delta_f,\delta_r]^{\mathrm T}$. Their admissible region is defined by
\begin{align}
G_{\mathrm{actr}}u&\preceq h_{\mathrm{actr}},
\quad u\in\mathcal C_{\mathrm{actr}},
\label{eq:actuator constraints}\\
G_{\mathrm{actr}}&=\begin{bmatrix}1&0\\0&1\\-1&0\\0&-1\end{bmatrix},
\quad
h_{\mathrm{actr}}=\begin{bmatrix}
\delta_{f,\max}\\\delta_{r,\max}\\
-\delta_{f,\min}\\-\delta_{r,\min}
\end{bmatrix}.
\nonumber
\end{align}
Here, $\delta_{a,\min}$ and $\delta_{a,\max}$ are the lower and upper steering-angle limits for axle $a\in\{f,r\}$. The symbol $\preceq$ denotes componentwise inequalities. The steering-rate limits are imposed between successive projected commands as
\begin{align}
&\Delta t\,\dot\delta_{\min}
\preceq u_{j|t}-u_{j-1|t}
\preceq\Delta t\,\dot\delta_{\max},
\label{eq:steering_rate_constraints}\\
&\dot\delta_{\max}=
[\dot\delta_{f,\max},\dot\delta_{r,\max}]^{\mathrm T},
\nonumber\\
&\dot\delta_{\min}=
[\dot\delta_{f,\min},\dot\delta_{r,\min}]^{\mathrm T}.
\nonumber
\end{align}
Here, $\Delta t$ is the prediction sampling interval, $j$ is the prediction-step index, and $t$ is the current sampling instant. At $j=0$, $u_{-1|t}=u_{t-1}$ denotes the previously applied steering command. At subsequent steps, $u_{j-1|t}$ denotes the preceding projected command. The actuator bounds used in the experiments are symmetric around zero.

\subsubsection{Constraints in State--Control Coupled Space}
Tire--road friction limitations couple vehicle states and steering inputs through the lateral tire forces. Define $\mathcal W=\{FL,FR,RL,RR\}$ as the wheel set and $a(w)\in\{f,r\}$ as the axle of wheel $w$. The kinematic slip terms are
\begin{align}
\eta_f(x)&=\frac{v_y+l_f\gamma}{v_x},
\quad
\eta_r(x)=\frac{v_y-l_r\gamma}{v_x}.
\label{eq:slip_kinematic_terms}
\end{align}
The linear per-wheel lateral tire force is approximated by
\begin{align}
F_{y,w}(x,u)
=C_{a(w)}[\delta_{a(w)}-\eta_{a(w)}(x)],
\label{eq:per_wheel_lateral_force}
\end{align}
where $C_{a(w)}$ is the cornering stiffness per tire and $\delta_{a(w)}$ is the corresponding axle steering angle. The normalized friction constraint is
\begin{align}
b_{\mathrm{tire},w}(x,u)
={}&1-\rho_{\mathrm{tire}}
-\left(\frac{F_{x,w}}{\mu F_{z,w}}\right)^2
\nonumber\\
&-\left(\frac{F_{y,w}(x,u)}{\mu F_{z,w}}\right)^2\geq0,
\label{eq:barrier function of tire ellipse}
\end{align}
where $F_{x,w}$ and $F_{z,w}$ are the longitudinal and normal wheel forces, $\mu$ is the tire--road friction coefficient, and $\rho_{\mathrm{tire}}$ is the friction reserve. Estimated quantities are indicated by a hat. For the current state and wheel-force estimates, the remaining lateral-force bound is evaluated as
\begin{align}
\bar F_{y,w}
={}&\hat\mu\hat F_{z,w}\sqrt{\max\Bigg\{
\varepsilon_{\mathrm{rad}},
1-\rho_{\mathrm{tire}}
-\left(\frac{\hat F_{x,w}}{\hat\mu\hat F_{z,w}}\right)^2
\Bigg\}},
\label{eq:tire_remaining_lateral_force}
\end{align}
where $\varepsilon_{\mathrm{rad}}=10^{-4}$ is the numerical lower bound used by the QP implementation. Normal loads and longitudinal speeds are also subject to positive numerical lower bounds. If the unmodified radicand is negative, the assumed longitudinal-force demand has already exceeded the reserved friction capacity. The numerical lower bound does not restore physical tire feasibility.

Each wheel contributes two control-affine inequalities:
\begin{align}
C_{a(w)}\delta_{a(w)}-s_{w,+}
&\leq\bar F_{y,w}+C_{a(w)}\eta_{a(w)},
\nonumber\\
-C_{a(w)}\delta_{a(w)}-s_{w,-}
&\leq\bar F_{y,w}-C_{a(w)}\eta_{a(w)}.
\label{eq:tire_wheel_inequalities}
\end{align}
Here, $s_{w,+}$ and $s_{w,-}$ are nonnegative relaxation variables. Stacking the inequalities for the four wheels gives
\begin{align}
G_{\mathrm{tire}}u-s_{\mathrm{tire}}
&\preceq h_{\mathrm{tire}},
\label{eq:tire adhesion ellipse linear constraint}
\end{align}
where $G_{\mathrm{tire}}\in\mathbb R^{8\times2}$, $h_{\mathrm{tire}}\in\mathbb R^8$, and $s_{\mathrm{tire}}\in\mathbb R_{\geq0}^8$. The row coefficients follow \eqref{eq:tire_wheel_inequalities}. This eight-inequality formulation retains the individual wheel-force limits and their associated slack penalties. The use of per-wheel constraints is consistent with the eight tire slack variables in the differentiable QP.

\subsubsection{Constraints in State Space}
The handling envelope constrains the lateral--yaw state and is mapped into steering-input inequalities through the vehicle dynamics. For $x=[v_y,\gamma]^{\mathrm T}$, a diamond-shaped envelope provides four planar state inequalities:
% 仅作用在状态空间的稳定性约束的通用数学表示为$b(x) \leq 0$, 典型的例子是基于相平面的稳定域约束。在安全滤波器优化问题的定义中，仅有控制变量为决策变量，状态为固定的输入，因此需要将此类约束映射到控制空间。基于菱形稳定域的质心侧偏角-横摆角速度约束为

\begin{align}
    &H_{handl} x \preceq \mathbf1_4 \label{eq:handling envelope original constraint} \\
    &H_{handl} = \begin{bmatrix}
        \frac{1}{\hat{v}_x \beta_{max}} & -\frac{1}{\hat{v}_x \beta_{max}} & -\frac{1}{\hat{v}_x \beta_{max}} & \frac{1}{\hat{v}_x \beta_{max}} \\
        \frac{1}{\gamma_{max}} & \frac{1}{\gamma_{max}} & -\frac{1}{\gamma_{max}} & -\frac{1}{\gamma_{max}}
    \end{bmatrix}^T \nonumber 
\end{align}
where $\beta_{\max}$ and $\gamma_{\max}$ denote the prescribed sideslip and yaw-rate boundaries, respectively. The longitudinal speed $\hat v_x$ is frozen during each local QP evaluation, and $\mathbf1_4$ denotes the four-dimensional vector of ones. The matrix $H_{\mathrm{handl}}\in\mathbb R^{4\times2}$ defines the handling boundary through the approximation $\beta\approx v_y/\hat v_x$. The four handling-envelope inequalities are
converted into control-dependent conditions
using relative-degree-one CBFs.
Forward invariance can be established for
the nominal continuous-time model when
the CBF inequalities are feasible and
enforced without relaxation. Since the relative degree of the handling envelope is one, the matrix form of the control barrier function based on the handling envelope is represented as:

\begin{align}
    &\dot{B}_{handl}(x,u) + \Lambda B_{handl}(x) \succeq \textbf{0}_4 \label{eq:cbf handling envelope differential} \\
    &\Lambda = \mathrm{diag}(\alpha_1, \ldots, \alpha_4) \nonumber \\
    &B_{handl}(x) = \begin{bmatrix}
        1 - \frac{v_y}{\hat{v}_x \beta_{max}} - \frac{\gamma}{\gamma_{max}} \\
        1 + \frac{v_y}{\hat{v}_x \beta_{max}} - \frac{\gamma}{\gamma_{max}} \\
        1 + \frac{v_y}{\hat{v}_x \beta_{max}} + \frac{\gamma}{\gamma_{max}} \\
        1 - \frac{v_y}{\hat{v}_x \beta_{max}} + \frac{\gamma}{\gamma_{max}}
    \end{bmatrix} \nonumber
\end{align}
Each $\alpha_i>0$ determines the permitted
decrease rate of the corresponding barrier
function. Its value influences the
state-dependent admissible control region.
A larger gain permits a faster decrease
of a positive barrier function and therefore
does not necessarily impose a more
conservative safety condition. $\dot{B}_{handl}(x,u)$ can be computed by substituting the vehicle dynamics model ~\eqref{eq:vehicle model} into the constraint. Rewrite this constraint into the unified control-affine form:

\begin{align}
&G_{\mathrm{handl}}u-s_{\mathrm{handl}}
\preceq h_{\mathrm{handl}},
\label{eq:cbf based on handling envelope}\\
&G_{\mathrm{handl}}
=H_{\mathrm{handl}}\mathcal G,
\nonumber\\
&h_{\mathrm{handl}}
=-H_{\mathrm{handl}}\mathcal F(x)
+\Lambda
(\mathbf 1_4-H_{\mathrm{handl}}x).
\nonumber
\end{align}
Here, $s_{\mathrm{handl}}\in\mathbb R_{\geq0}^{4}$ denotes the handling-envelope slack vector. The matrix $\Lambda=\operatorname{diag}(\alpha_1,\ldots,\alpha_4)$ contains positive CBF gains. The nominal handling set is $\mathcal C_{\mathrm{handl}}=\{x:H_{\mathrm{handl}}x\preceq\mathbf1_4\}$. The formulation evaluates the handling envelope using the operating parameters available at each QP step. These parameters are held fixed during the local CBF calculation. If the envelope matrix varies continuously with time, its derivative must also be considered. The corresponding constraint contains the additional term $-\dot H_{\mathrm{handl}}x$ on the right-hand side.

The slack variables relax the nominal CBF inequalities when required by the QP. Consequently, the resulting soft-constrained filter does not establish strict forward invariance of the original handling envelope.

\subsection{Chassis-Constrained QP Formulation}\label{sec:Construction of Unified Safety Filter for Vehicle Stability Control}
The preceding constraints provide a unified control-space representation of steering- actuator, tire-force, and handling-stability limitations. To account for their evolution along a predicted vehicle trajectory, a sequence of state-dependent safety-projection problems is constructed over a horizon of $N$ steps. The local prediction horizon $N$ equals the steering-policy output horizon $H_c=10$ in the reported model.

At sampling instant $t$, the nominal policy provides the steering sequence
\begin{align}
\mathbf U_t^{\mathrm{ref}}
=\big[u_{0|t}^{\mathrm{ref}},\ldots,
u_{N-1|t}^{\mathrm{ref}}\big]^{\mathrm T}
\in\mathbb R^{N\times2}.
\end{align}
The initial predicted state is $\hat x_{0|t}=\hat x_t$. At each prediction step $j$, the nominal steering command is corrected using a minimum-intervention quadratic program.

The optimization variables are
\begin{align}
z_{j|t}
=
\operatorname{col}
\left(
u_{j|t},
s_{\mathrm{handl},j|t},
s_{\mathrm{tire},j|t}
\right),
\end{align}
where $s_{\mathrm{handl},j|t}$ and $s_{\mathrm{tire},j|t}$ represent the relaxation variables of the handling-envelope and tire-friction constraints.

The local safety-projection problem is
formulated as
\begin{align}
z_{j|t}^{*}
={}&
\arg\min_{z_{j|t}}
\mathcal J_{j|t},
\label{eq:augmented_qp_physical}\\
\mathcal J_{j|t}
={}&
\left\|
u_{j|t}-u_{j|t}^{\mathrm{ref}}
\right\|_{Q_{u,j}}^2
\nonumber\\
&+
\left\|
s_{\mathrm{handl},j|t}
\right\|_{W_{\mathrm{handl}}}^2
\nonumber\\
&+
\left\|
s_{\mathrm{tire},j|t}
\right\|_{W_{\mathrm{tire}}}^2,
\nonumber\\
\mathrm{s.t.}\quad
&G_{\mathrm{actr}}u_{j|t}
\preceq h_{\mathrm{actr}},
\nonumber\\
&\Delta t\dot\delta_{\min}
\preceq u_{j|t}-u_{j-1|t}
\preceq\Delta t\dot\delta_{\max},
\nonumber\\
&G_{\mathrm{handl}}(\hat x_{j|t})u_{j|t}
-s_{\mathrm{handl},j|t}
\nonumber\\
&\qquad\preceq
h_{\mathrm{handl}}
(\hat x_{j|t},\alpha_{j|t}),
\nonumber\\
&G_{\mathrm{tire}}(\hat x_{j|t})u_{j|t}
-s_{\mathrm{tire},j|t}
\nonumber\\
&\qquad\preceq
h_{\mathrm{tire}}
(\hat x_{j|t},\rho_{j|t}),
\nonumber\\
&s_{\mathrm{handl},j|t}\succeq0,
\quad
s_{\mathrm{tire},j|t}\succeq0.
\nonumber
\end{align}

The positive-definite matrix $Q_{u,j}$ determines the relative penalty assigned to front- and rear-wheel steering corrections. The positive-definite slack-weight matrices $W_{\mathrm{handl}}\in\mathbb R^{4\times4}$ and $W_{\mathrm{tire}}\in\mathbb R^{8\times8}$ penalize violations of the corresponding nominal constraints. In the implementation, their diagonal penalty coefficients are both $10^4$. The vector $\alpha_{j|t}\in\mathbb R_{>0}^4$ and scalar $\rho_{j|t}$ specify the handling-CBF gains and tire-friction reserve at prediction step $j$. The QP decision vector has dimension 14, comprising two steering angles, four handling slacks, and eight tire slacks.

After solving the local QP, the optimized steering command is extracted as
\begin{align}
u_{j|t}^{*}
=
[z_{j|t}^{*}]_u.
\end{align}
The predicted lateral--yaw state is
subsequently updated using
\begin{align}
\hat x_{j+1|t}
={}&
\hat x_{j|t}
+\Delta t
\Big[
\mathcal F(\hat x_{j|t})
+\mathcal G u_{j|t}^{*}
\Big].
\label{eq:projected_state_rollout}
\end{align}

The updated state determines the tire-force and handling-envelope constraints used at the next prediction step. This procedure is repeated for $j=0,\ldots, N-1$.

The resulting projected steering sequence is
\begin{align}
\mathbf U_t^{\mathrm{opt}}
=\big[u_{0|t}^{*},\ldots,u_{N-1|t}^{*}\big]^{\mathrm T}
\in\mathbb R^{N\times2}.
\label{eq:projected_control_sequence}
\end{align}

The formulation consists of sequential QP solutions linked by vehicle-dynamics propagation. Each QP minimizes the local steering correction under the predicted state and the corresponding constraints. The projected commands also influence later optimization problems through their effects on the predicted states.

During receding-horizon execution, the first projected command is applied when safety-filtered control is selected. The remaining projected commands and predicted states provide temporal information for offline policy training.

\subsection{Dynamics-Unrolled Differentiable Projection}

Conventional safety filters with fixed model parameters and prescribed safety margins may introduce inaccurate or overly conservative control constraints under vehicle-model uncertainty and state-estimation errors. To address this limitation, the preceding physical constraints are embedded into a differentiable quadratic programming (QP) layer. The projection weights, CBF gains, and tire-friction reserve are parameterized to support data-driven adaptation of the admissible control region under varying operating conditions.

Combining the sequential QPs with vehicle-state propagation yields
\begin{align}
\mathbf U_t^{\mathrm{opt}}
=
\mathcal D_{\mathrm{PSF}}
\left(
\mathbf U_t^{\mathrm{ref}},
p_t^{\mathrm{seq}},
\hat x_t,
u_{t-1}
\right),
\label{eq:differentiable_psf_operator}
\end{align}
where $\mathcal D_{\mathrm{PSF}}$ denotes the differentiable safety-projection operator, $\mathbf U_t^{\mathrm{ref}}$ and $\mathbf U_t^{\mathrm{opt}}$ are the nominal and projected steering sequences, $p_t^{\mathrm{seq}}$ contains the projection weights and safety-related parameters, $\hat x_t$ is the estimated vehicle state, and $u_{t-1}$ is the previously applied steering command. Under appropriate regularity conditions, QP solution sensitivities can be obtained by differentiating the Karush--Kuhn--Tucker (KKT) conditions \cite{Gould_2021,amos2021optnetdifferentiableoptimizationlayer}. This enables learning-based adjustment of the safety-projection parameters and provides the basis for the policy--safety coordination framework developed in the following section.

\section{Safety-Guarded Task Learning}
\label{sec:Safety-Guaranteed Bi-Level Policy Optimization Framework}
This section presents the proposed learning framework, including its network architecture, differentiable policy--safety coupling, and two-stage training strategy. The framework integrates nominal steering prediction, safety-parameter generation, and differentiable handling-stability projection to coordinate trajectory-tracking performance with physical constraints.

% \begin{figure}[!t]
%     \centering
%     \includegraphics[width=0.8\textwidth]{model_framework0729.jpg}
%     \caption{Architecture of the proposed policy--safety learning framework. The policy and safety-parameter networks generate nominal steering commands and projection parameters from vehicle observations. The differentiable safety-projection layer produces corrected steering commands and provides optimization sensitivities for safety-guided learning.}
%     \label{fig:model_framework}
% \end{figure}

\begin{figure}[!t]
    \centering
    \includegraphics[width=\columnwidth]{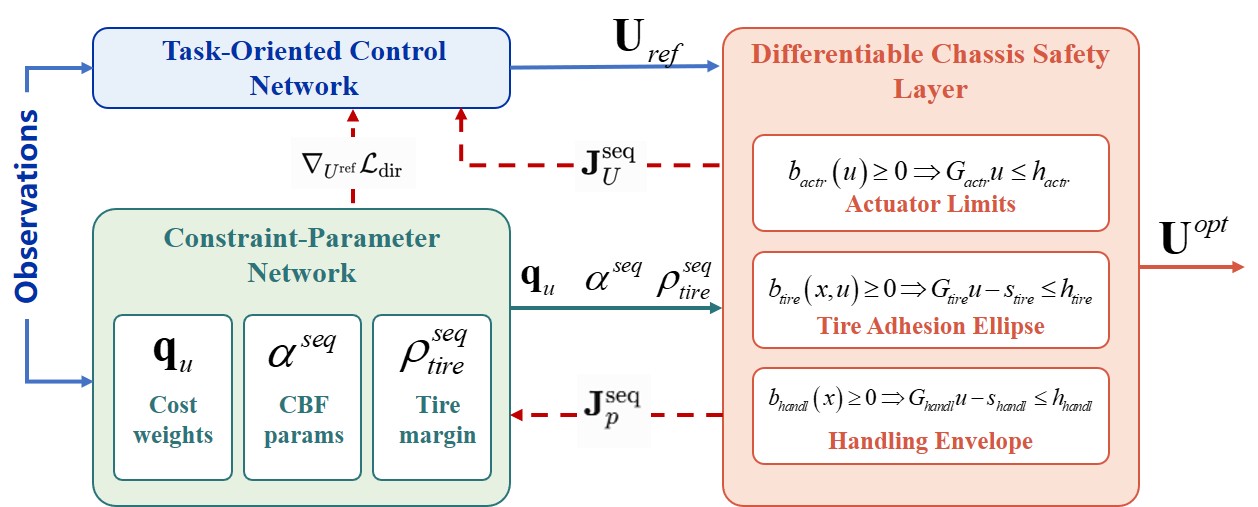}
    \caption{Architecture of the proposed safety-guarded task learning framework.}
    \label{fig:model_framework}
\end{figure}

\subsection{Modular Architecture and Signal Flow}

As shown in Fig.~\ref{fig:model_framework}, the framework consists of a policy network, a safety-parameter generation network, and a differentiable handling-stability projection layer. Both networks process vehicle observations through a shared feature representation to generate
\begin{align}
h_t &= \mathcal E_\eta(o_t),
\nonumber\\
\mathbf U_t^{\mathrm{ref}}
&= \mathcal D_{u,\phi}(h_t),
\nonumber\\
p_t^{\mathrm{seq}}
&= \mathcal D_{p,\psi}(h_t),
\label{eq:policy_parameter_generation}
\end{align}
where $o_t$ denotes the observation vector, $h_t$ is the encoded feature representation, and $\eta$, $\phi$, and $\psi$ are the corresponding network parameters. The nominal steering sequence $\mathbf U_t^{\mathrm{ref}}\in\mathbb R^{H_c\times2}$ contains the predicted front- and rear-wheel steering commands over $H_c$ steps. The safety-parameter sequence $p_t^{\mathrm{seq}}$ comprises the projection weights $\mathbf q_{u,t}$, handling-CBF gains $\boldsymbol\alpha_t^{\mathrm{seq}}$, and tire-friction reserves $\boldsymbol\rho_{\mathrm{tire},t}^{\mathrm{seq}}$.

The differentiable projection layer receives $\mathbf U_t^{\mathrm{ref}}$ and $p_t^{\mathrm{seq}}$ to generate the corrected sequence $\mathbf U_t^{\mathrm{opt}}$ under the prescribed physical constraints. During training, task-related losses and projection sensitivities provide gradient feedback to the upstream networks, subject to the enabled loss terms and trainable parameter groups. The corresponding gradient propagation and optimization strategy are detailed in the following subsections.

\subsection{Projection-Guided Gradient Coupling}
\label{subsec:implicit_gradient}

ChassisOptNet couples task-oriented learning with physical constraint optimization through the differentiable projection defined in \eqref{eq:differentiable_psf_operator}. As illustrated in Fig.~\ref{fig:model_framework}, the nominal control sequence and constraint parameters are jointly involved in a dynamics-unrolled QP process, allowing projection-dependent losses to influence network optimization.

Let $\mathbf Z_t^*$ collect the projected control sequence and optimized constraint slacks. The gradients with respect to the nominal commands and projection parameters are
\begin{align}
\nabla_{\mathbf U_t^{\mathrm{ref}}}\mathcal L
={}&
\nabla_{\mathbf U_t^{\mathrm{ref}}}\mathcal L_{\mathrm{dir}}
+
(\mathbf J_U^{\mathrm{seq}})^{\mathrm T}
\nabla_{\mathbf Z_t^*}\mathcal L_{\mathrm{proj}},
\nonumber\\
\nabla_{p_t^{\mathrm{seq}}}\mathcal L
={}&
(\mathbf J_p^{\mathrm{seq}})^{\mathrm T}
\nabla_{\mathbf Z_t^*}\mathcal L_{\mathrm{proj}},
\label{eq:implicit coupled gradient}
\end{align}
where $\mathcal L_{\mathrm{dir}}$ denotes direct task and risk supervision, while $\mathcal L_{\mathrm{proj}}$ represents projection-dependent losses. The Jacobians are defined as
$\mathbf J_U^{\mathrm{seq}}
=\partial\mathbf Z_t^*/\partial\mathbf U_t^{\mathrm{ref}}$
and
$\mathbf J_p^{\mathrm{seq}}
=\partial\mathbf Z_t^*/\partial p_t^{\mathrm{seq}}$.

Unlike independent single-step projections, these Jacobians capture both the local QP sensitivities and the effects of preceding control corrections on subsequent vehicle states and constraints. The projection weights determine the control-correction metric, whereas the CBF gains and tire-friction margins parameterize the prescribed constraints. This establishes a gradient coupling between task-oriented control and chassis constraint projection.

During training, nominal-rollout risk losses and detached intervention targets provide direct policy guidance, while projection-dependent losses enable gradients through the differentiable QPs. The latter also support optimization of the constraint-parameter network when its parameters are trainable. Consequently, the effective gradient coupling depends on the enabled loss terms and parameter-update configuration.

\subsection{Two-Stage Task–Constraint Learning}
\label{subsec:two_stage_training}

The proposed framework adopts a two-stage training strategy. Stage~I initializes the nominal steering policy using expert MPC demonstrations. Stage~II introduces handling-risk supervision and differentiable safety projection to improve the coordination between tracking performance and physical constraints.

\subsubsection{Task-Oriented Learning}

The nominal policy is trained to predict the expert steering sequence. Let $\mathbf U_n^{\mathrm{ref}}$ and $\mathbf U_n^{\mathrm{exp}}\in\mathbb R^{H_c\times2}$ denote the nominal and expert steering sequences of sample $n$, where $H_c$ is the prediction horizon and $B$ is the batch size.

To account for prediction-step importance, define the weighted mean-square operator
\begin{align}
\mathcal M_J(a)
&=\frac{1}{BJd_a}
\sum_{n=1}^{B}\sum_{j=0}^{J-1}
w_j^{(J)}\|a_{n,j}\|_2^2,
\nonumber\\
w_j^{(J)}
&=\frac{Jd_h^j}
{\sum_{k=0}^{J-1}d_h^k},
\label{eq:weighted_mse}
\end{align}
where $J$ is the sequence length, $d_a$ is the dimension of $a_{n,j}$, and $d_h$ is the temporal decay factor.

The Stage-I objective is
\begin{align}
\mathcal L_{\mathrm I}
&=\mathcal L_{\mathrm{nom}}
+\lambda_{\mathrm{sm,I}}\mathcal L_{\mathrm{sm}},
\label{eq:stage1_loss}\\
\mathcal L_{\mathrm{nom}}
&=\mathcal M_{H_c}
(\mathbf U^{\mathrm{ref}}-\mathbf U^{\mathrm{exp}}),
\nonumber\\
\mathcal L_{\mathrm{sm}}
&=\frac{1}{2B(H_c-1)}
\sum_{n=1}^{B}\sum_{j=0}^{H_c-2}
\|u_{n,j+1}^{\mathrm{ref}}
-u_{n,j}^{\mathrm{ref}}\|_2^2.
\nonumber
\end{align}
Here, $\mathcal L_{\mathrm{nom}}$ provides expert steering supervision, while $\mathcal L_{\mathrm{sm}}$ suppresses abrupt variations in the predicted steering sequence. The safety-projection layer is not activated during Stage~I.

\subsubsection{Risk-Guided Fine-Tuning}
\label{subsubsec:stage2}

Stage~II initializes the policy from Stage~I and introduces safety-related learning objectives. The loss combines expert imitation, risk-gated intervention, expert-relative handling-risk supervision, curvature-dependent tracking, and QP slack regularization:
\begin{align}
\mathcal L_{\mathrm{II}}
={}&
\lambda_{\mathrm{nom}}\mathcal L_{\mathrm{nom}}
+\lambda_0\mathcal L_0
+\lambda_{\mathrm{sm}}\mathcal L_{\mathrm{sm}}
\nonumber\\
&+\lambda_{\mathrm{slk}}\mathcal L_{\mathrm{slk}}
+r_e\Big(
\lambda_{\mathrm{int}}\mathcal L_{\mathrm{int}}
+\lambda_{\mathrm{fint}}\mathcal L_{\mathrm{fint}}
\nonumber\\
&\quad+
\lambda_{\mathrm{rec}}\mathcal L_{\mathrm{rec}}
+\lambda_\kappa\mathcal L_\kappa
+\lambda_{\mathrm{term}}\mathcal L_{\mathrm{term}}
\Big),
\label{eq:stage2_total_loss}
\end{align}
where $\lambda$ denotes the corresponding loss weight and
$r_e=\min(1,e/E_{\mathrm{ramp}})$ gradually introduces the auxiliary objectives during the first $E_{\mathrm{ramp}}$ Stage-II epochs.

\noindent\textit{1) Expert imitation.}
To preserve the tracking performance learned in Stage~I, nominal-sequence imitation is retained. An additional first-step loss emphasizes the command used in receding-horizon execution:
\begin{align}
\mathcal L_0
=\frac{1}{2B}\sum_{n=1}^{B}
\|u_{n,0}^{\mathrm{ref}}
-u_{n,0}^{\mathrm{exp}}\|_2^2.
\label{eq:first_step_loss}
\end{align}

\noindent\textit{2) Risk-gated intervention.}
The handling-envelope utilization $P$ and maximum tire-force utilization $T$ are evaluated through nominal and expert vehicle-dynamics rollouts:
\begin{align}
P_{n,j}
&=\frac{|\beta_{n,j}|}{\beta_{\max}}
+\frac{|\gamma_{n,j}|}{\gamma_{\max,n,j}},
\nonumber\\
T_{n,j}
&=\max_{w\in\mathcal W}
\frac{
\sqrt{F_{x,w,n}^{2}+F_{y,w,n,j}^{2}+\varepsilon}
}{
\mu_nF_{z,w,n}+\varepsilon
}.
\label{eq:training_risk_indices}
\end{align}
Here, $\beta$ and $\gamma$ denote the predicted sideslip angle and yaw rate, $\mathcal W$ is the four-wheel set, and $\varepsilon$ is a numerical regularizer. The limits $\beta_{\max}$ and $\gamma_{\max,n,j}$ follow the handling-envelope definition in Section~\ref{sec:idcbf}. The quantities $F_x$, $F_y$, $F_z$, and $\mu$ denote longitudinal tire force, lateral tire force, vertical load, and road adhesion coefficient, respectively.

Let superscripts $\mathrm{ref}$ and $\mathrm{exp}$ identify nominal-policy and expert rollouts. The expert-relative risk allowances are
\begin{align}
\bar P_{n,j}
&=\max(1+\varepsilon_r,
P_{n,j}^{\mathrm{exp}}+m_r),
\nonumber\\
\bar T_{n,j}
&=\max(1+\varepsilon_r,
T_{n,j}^{\mathrm{exp}}+m_r),
\label{eq:expert_relative_risk_thresholds}
\end{align}
where $\varepsilon_r$ is the transient risk tolerance and $m_r$ is the expert-relative margin.

The intervention gate and corresponding losses are
\begin{align}
g_{n,j}
&=\mathbb I\left[
P_{n,j}^{\mathrm{ref}}>\bar P_{n,j}
\ \lor\
T_{n,j}^{\mathrm{ref}}>\bar T_{n,j}
\right],
\nonumber\\
d_{n,j}
&=g_{n,j}
\left(
u_{n,j}^{\mathrm{ref}}
-\operatorname{sg}(u_{n,j}^{\mathrm{opt}})
\right),
\nonumber\\
\mathcal L_{\mathrm{int}}
&=\mathcal M_{H_c}(d),
\nonumber\\
\mathcal L_{\mathrm{fint}}
&=\mathcal M_{H_c-1}
\left(\{d_{n,j}\}_{j=1}^{H_c-1}\right),
\label{eq:stage2_intervention_losses}
\end{align}
where $\mathbb I[\cdot]$ is the indicator function and $\operatorname{sg}$ denotes stop-gradient. Both the gate and projected steering targets are detached during differentiation. These losses guide the nominal policy toward feasible steering decisions when its predicted risk exceeds the expert-relative allowance.

\begin{table}[!t]
\centering
\caption{Key network and training hyperparameters.}
\label{tab:training_loss_parameters}
\footnotesize
\setlength{\tabcolsep}{4pt}
\renewcommand{\arraystretch}{1.08}
\begin{tabularx}{\columnwidth}{@{}Xc@{}}
\toprule
Parameter & Value\\
\midrule
\multicolumn{2}{@{}l}{\textit{Network Architecture}}\\
Input dimension & $22+4H$\\
Preview/control horizon $(H,H_c)$ & $(10,10)$\\
Hidden dimension & $192$\\
Residual blocks & $3$\\
Dropout rate & $0.05$\\
Output dimension & $H_c\times(2+7)$\\
\midrule
\multicolumn{2}{@{}l}{\textit{Training Configuration}}\\
Optimizer & AdamW\\
Batch size & $128$\\
Stage-I / Stage-II epochs & $160/340$\\
Stage-I / Stage-II learning rates
& $10^{-3}/2\times10^{-4}$\\
Curriculum ramp epochs & $30$\\
\midrule
\multicolumn{2}{@{}l}{\textit{Key Stage-II Loss Weights}}\\
$(\lambda_{\mathrm{nom}},\lambda_0)$
& $(1.0,\,2.0)$\\
$(\lambda_{\mathrm{int}},\lambda_{\mathrm{fint}})$
& $(0.03,\,0.05)$\\
$\lambda_{\mathrm{rec}}$ & $0.02$\\
$(\lambda_\kappa,\lambda_{\mathrm{term}})$
& $(0.02,\,0.005)$\\
$\lambda_{\mathrm{sm}}$ & $10^{-4}$\\
$\lambda_{\mathrm{slk}}$ & $10^{-7}$\\
\bottomrule
\end{tabularx}
\end{table}

\noindent\textit{3) Expert-relative risk supervision.}
To discourage excessive risk without suppressing successful near-limit responses, a recoverability-related penalty is constructed from handling and tire-risk sequences:
\begin{align}
\mathcal L_{\mathrm{rec}}
=\sum_{c\in\{P,T\}}
\big(
\mathcal L_{\mathrm{exc}}^c
+\mathcal L_{\mathrm{bud}}^c
+\mathcal L_{\mathrm{peak}}^c
+\mathcal L_{\mathrm{trend}}^c
\big).
\label{eq:stage2_recoverability_loss}
\end{align}
For risk category $c$, the four components penalize the squared excess above the expert-relative allowance, additional mean boundary violation beyond budget $b_r$, excessive peak risk, and unfavorable risk growth relative to the expert, respectively. The trend penalty uses the difference between successive risk indices and coefficient $k_r$. The expert risk trajectories are treated as fixed references. This formulation encourages bounded risk excursions and recovery trends without imposing a terminal invariant-set condition.

\noindent\textit{4) Curvature-dependent tracking.}
For high-curvature maneuvers, the desired yaw rate is $\gamma_{n,j}^{d}=v_{x,n}\kappa_{n,j}$. A curvature-dependent gate is defined as
\begin{align}
c_n
=\sigma\left(
\frac{\max_j|\kappa_{n,j}|-\kappa_0}
{\sigma_\kappa}
\right),
\label{eq:stage2_curvature_gate}
\end{align}
where $\sigma(\cdot)$ is the sigmoid function, $\kappa_{n,j}$ is the preview curvature, and $\kappa_0$ and $\sigma_\kappa$ determine the activation threshold and transition width.

The tracking and terminal penalties are
\begin{align}
\mathcal L_\kappa
&=\mathcal M_{H_c}
\left(
c_n[\hat\gamma_{n,j}^{\mathrm{ref}}
-\gamma_{n,j}^{d}]
\right),
\nonumber\\
\mathcal L_{\mathrm{term}}
&=\frac{1}{B}\sum_{n=1}^{B}c_n
\Big[
(\hat e_{y,n,H_c-1}^{\mathrm{ref}})^2
\nonumber\\
&\quad+
4(\hat e_{\psi,n,H_c-1}^{\mathrm{ref}})^2
\nonumber\\
&\quad+
0.5(\hat\gamma_{n,H_c-1}^{\mathrm{ref}}
-\gamma_{n,H_c-1}^{d})^2
\Big],
\label{eq:stage2_curvature_tracking}
\end{align}
where $\hat e_y^{\mathrm{ref}}$, $\hat e_\psi^{\mathrm{ref}}$, and $\hat\gamma^{\mathrm{ref}}$ are the predicted lateral error, heading error, and yaw rate under the nominal sequence. These terms improve anticipatory yaw response and terminal tracking during sustained high-curvature maneuvers.

\noindent\textit{5) Constraint regularization.}
The QP slack loss is
\begin{align}
\mathcal L_{\mathrm{slk}}
={}&\frac{1}{BH_c}\sum_{n,j}
\Big(
\frac{\|s_{\mathrm{handl},n,j}^{*}\|_2^2}{4}
\nonumber\\
&\qquad+
\frac{\|s_{\mathrm{tire},n,j}^{*}\|_2^2}{8}
\Big),
\label{eq:stage2_slack_loss}
\end{align}
where $s_{\mathrm{handl}}^*\in\mathbb R^4$ and $s_{\mathrm{tire}}^*\in\mathbb R^8$ are the optimized handling and tire-constraint slack vectors. This term discourages constraint relaxation and retains a gradient pathway through the differentiable QP.

The policy and shared feature layers are updated during Stage~II. The safety-parameter output layer is optimized only when its training option is enabled. In the reference configuration, projected-sequence imitation and optional nominal-rollout tracking have zero loss weights, while the safety-parameter output layer remains fixed.

\section{Experimental Validation}
\label{sec:Training and Online Testing Experiments}

The proposed framework is evaluated using closed-loop Python--CarSim simulations. The experiments include generalization tests under varying trajectories, speeds, and road adhesion, a near-limit handling maneuver, and ablation studies of the safety-projection and learning components.

\subsection{Simulation Platform and Expert Demonstrations}

Expert demonstrations were collected using a nonlinear model predictive controller (MPC) in a Python--CarSim co-simulation environment. The MPC employed a single-track lateral--yaw model and nonlinear Frenet-frame kinematics to predict vehicle motion. The prediction state and control input were defined as
$\boldsymbol\chi=[s,e_d,e_\psi,v_x,v_y,\gamma]^{\mathrm T}$
and
$\mathbf u=[\delta_f,\delta_r]^{\mathrm T}$,
where $s$ is the path coordinate, $e_d$ and $e_\psi$ are the lateral and heading errors, $v_x$ and $v_y$ are the longitudinal and lateral velocities, and $\gamma$ is the yaw rate. The MPC optimized the steering sequence by minimizing
\begin{align}
J_{\mathrm{MPC}}
={}&
\sum_{j=1}^{N_p}
\mathbf y_j^{\mathrm T}\mathbf Q\mathbf y_j
+\sum_{j=0}^{N_p-1}
\mathbf u_j^{\mathrm T}\mathbf R\mathbf u_j
\nonumber\\
&+
\sum_{j=0}^{N_p-1}
\Delta\mathbf u_j^{\mathrm T}
\mathbf R_\Delta\Delta\mathbf u_j,
\label{eq:expert_mpc_training_cost}
\end{align}
where
$\mathbf y_j=[e_{d,j},\dot e_{d,j},e_{\psi,j},
\dot e_{\psi,j}]^{\mathrm T}$
and
$\Delta\mathbf u_j=(\mathbf u_j-\mathbf u_{j-1})/\Delta t$.
The weighting matrices were
$\mathbf Q=\operatorname{diag}(30,60,10,20)$
and
$\mathbf R=\mathbf R_\Delta
=\operatorname{diag}(10^{-2},10^{-2})$.
The sampling interval was $\Delta t=0.05~\mathrm{s}$, with a prediction horizon of $N_p=20$.
Steering-actuator, tire-friction, and handling-stability constraints were included in the optimization. The resulting nonlinear program was solved using CasADi and IPOPT, with warm-start initialization. Only the first optimized steering command was applied, while an independent PID controller regulated longitudinal speed.

\begin{table}[!t]
\caption{Vehicle and steering-actuator parameters.}
\label{tab:vehicle hyperparameters}
\centering
\footnotesize
\renewcommand{\arraystretch}{1.12}
\begin{tabular}{@{}lccc@{}}
\toprule
Parameter & Symbol & Value & Unit\\
\midrule
Vehicle mass & $m$ & 1259.9 & kg\\
Yaw moment of inertia & $I_z$ & 1343.1 & kg$\cdot$m$^2$\\
CG to front axle & $l_f$ & 1.04 & m\\
CG to rear axle & $l_r$ & 1.56 & m\\
Front cornering stiffness per tire & $C_f$ & 39800 & N/rad\\
Rear cornering stiffness per tire & $C_r$ & 34050 & N/rad\\
Effective wheel radius & $R_w$ & 0.287 & m\\
\bottomrule
\end{tabular}
\end{table}
\begin{table}[!t]
\centering
\caption{Statistical comparison of tracking accuracy and handling stability in trajectory, speed, and tire--road friction generalization tests.}
\label{tab:generalization_comparison}
\footnotesize
\setlength{\tabcolsep}{7pt}
\renewcommand{\arraystretch}{1.06}
\begin{tabular}{@{}p{0.54\linewidth}ccc@{}}
\toprule
Metric & LQR & MPC & Ours \\
\midrule
\addlinespace[4pt]
\multicolumn{4}{@{}l}{\textit{Trajectory generalization}}\\
\midrule
Maximum absolute lateral error (m) & 0.1482 & 0.1564 & \textbf{0.0878} \\
Mean absolute lateral error (m) & 0.0342 & 0.0358 & \textbf{0.0199} \\
Maximum absolute sideslip angle (deg) & 1.2078 & 1.1229 & 2.0003 \\
Mean absolute sideslip angle (deg) & 0.1852 & 0.1675 & 0.3732 \\
Maximum absolute yaw rate (deg/s) & 11.2793 & 11.6960 & 13.6922 \\
Mean absolute yaw rate (deg/s) & 2.1270 & 2.1417 & 2.3611 \\
\addlinespace[4pt]
\multicolumn{4}{@{}l}{\textit{Velocity generalization}}\\
\midrule
Maximum absolute lateral error (m) & 0.1882 & 0.2226 & \textbf{0.1783} \\
Mean absolute lateral error (m) & \textbf{0.0400} & 0.0509 & 0.0443 \\
Maximum absolute sideslip angle (deg) & 2.3823 & 1.8960 & 5.0198 \\
Mean absolute sideslip angle (deg) & 0.4269 & 0.3079 & 1.5111 \\
Maximum absolute yaw rate (deg/s) & 17.5158 & 17.6572 & 20.4647 \\
Mean absolute yaw rate (deg/s) & 3.3605 & 3.5145 & 3.9477 \\
\addlinespace[4pt]
\multicolumn{4}{@{}l}{\textit{Tire--road adhesion generalization}}\\
\midrule
Maximum absolute lateral error (m) & 0.1662 & 0.1862 & \textbf{0.1199} \\
Mean absolute lateral error (m) & 0.0363 & 0.0449 & \textbf{0.0231} \\
Maximum absolute sideslip angle (deg) & 1.9426 & 1.1362 & 2.7342 \\
Mean absolute sideslip angle (deg) & 0.3531 & 0.1865 & 0.4839 \\
Maximum absolute yaw rate (deg/s) & 10.7256 & 10.7493 & 13.1834 \\
Mean absolute yaw rate (deg/s) & 2.1133 & 2.1663 & 2.3393 \\
\bottomrule
\end{tabular}
\end{table}

\begin{figure}[!t]
\centering
\begin{subfigure}[t]{0.32\linewidth}
  \centering
  \includegraphics[width=\linewidth]{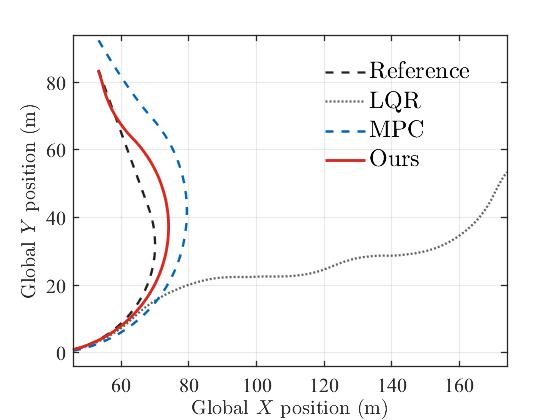}
  \caption{}\label{fig:flag104_traj}
\end{subfigure}\hfill
\begin{subfigure}[t]{0.32\linewidth}
  \centering
  \includegraphics[width=\linewidth]{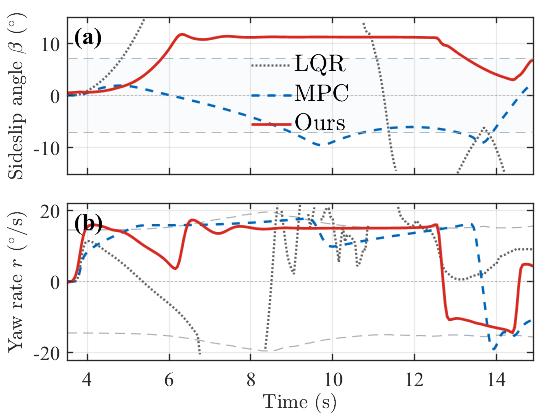}
  \caption{}\label{fig:flag104_beta_yawrate}
\end{subfigure}\hfill
\begin{subfigure}[t]{0.32\linewidth}
  \centering
  \includegraphics[width=\linewidth]{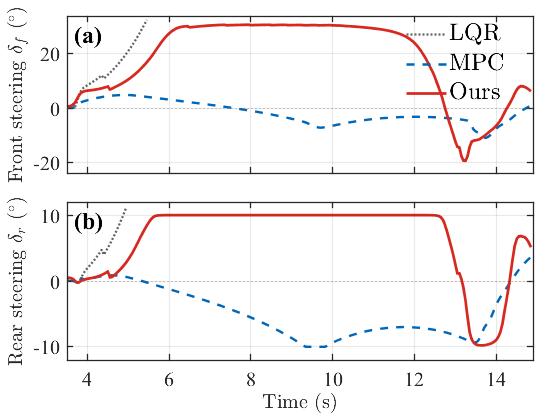}
  \caption{}\label{fig:flag104_steering}
\end{subfigure}

\medskip
\begin{subfigure}[t]{0.32\linewidth}
  \centering
  \includegraphics[width=\linewidth]{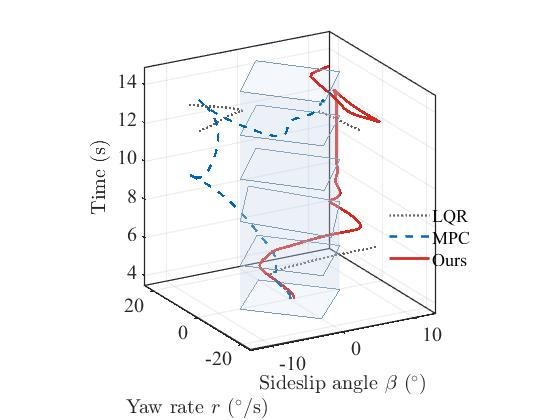}
  \caption{}\label{fig:flag104_3d}
\end{subfigure}\hfill
\begin{subfigure}[t]{0.32\linewidth}
  \centering
  \includegraphics[width=\linewidth]{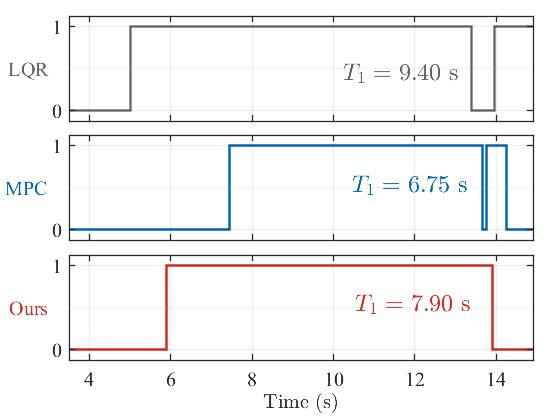}
  \caption{}\label{fig:flag104_boundary_exceedance}
\end{subfigure}\hfill
\begin{subfigure}[t]{0.32\linewidth}
  \centering
  \includegraphics[width=\linewidth]{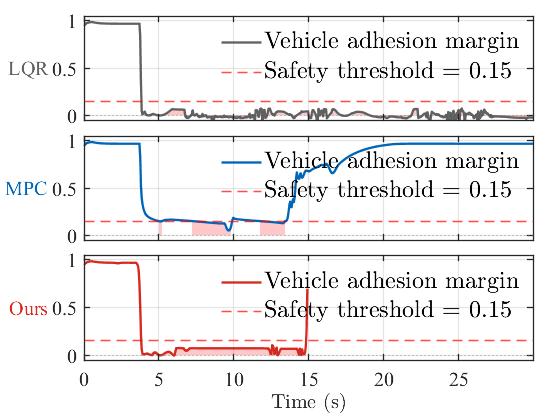}
  \caption{}\label{fig:flag104_tire_adhesion_safety_margin}
\end{subfigure}
\caption{Safety validation of the proposed controller under near-limit maneuvers: (a) trajectory-tracking result; (b) sideslip angle and yaw rate; (c) front and rear steering commands; (d) evolution of the dynamic stability envelope; (e) stability-boundary exceedance; (f) evolution of the tire-adhesion safety margin.}
\label{fig:flag104_test}
\end{figure} 
\begin{table}[!t]
\centering
\caption{Quantitative results of the ablation study under high-speed emergency collision-avoidance maneuvers.}
\label{tab:ablation_study}

\renewcommand{\arraystretch}{1.18}
\setlength{\tabcolsep}{4pt}
\footnotesize

\begin{tabularx}{\textwidth}
{@{}>{\raggedright\arraybackslash}p{0.18\textwidth}CCCC@{}}
\toprule
\textbf{Metric}
&
\textbf{Policy Network}
&
\textbf{Policy Network + Non-diff Safety Layer}
&
\textbf{Policy Network + Diff Safety Layer + No Joint Training}
&
\textbf{Ours}
\\
\midrule

Maximum Absolute Lateral Error (m)
& 0.4562
& --
& 1.8739
& \textbf{0.3422}
\\

Mean Absolute Lateral Error (m)
& 0.1150
& --
& 0.4981
& \textbf{0.1011}
\\

Maximum $|\beta|$ (deg)
& 7.9939
& --
& 9.5251
& 8.5579
\\

Mean $|\beta|$ (deg)
& 1.6558
& --
& 3.5399
& 2.4303
\\

Maximum $|\gamma|$ (deg/s)
& 23.6297
& --
& 14.0890
& 25.6503
\\

Mean $|\gamma|$ (deg/s)
& 4.7406
& --
& 4.3293
& 4.7711
\\

\bottomrule
\end{tabularx}

\vspace{1mm}

\begin{minipage}{\linewidth}
\footnotesize
\textit{Notes:}
Policy Network + Non-diff Safety Layer denotes the policy network followed by a fixed non-differentiable safety layer with $\alpha=5$ and $\rho_{\mathrm{tire}}=0.15$.
\end{minipage}

\end{table}
The expert dataset covered seven reference-path families, including lane changes, slalom, circular paths, J-turns, U-turns, and variable-curvature cornering. Reference speeds ranged from $30$ to $60~\mathrm{km/h}$, with tire--road adhesion coefficients between $0.6$ and $1.0$. These scenarios provided diverse steering demands and lateral--yaw responses.

% \begin{table}[!t]
% \centering
% \caption{Reference-path families used for expert demonstrations.}
% \label{tab:training_reference_maneuvers}
% \scriptsize
% \setlength{\tabcolsep}{2.5pt}
% \renewcommand{\arraystretch}{1.12}
% \begin{tabularx}{\columnwidth}{@{}
% >{\raggedright\arraybackslash}p{0.22\columnwidth}
% >{\raggedright\arraybackslash}X
% >{\raggedright\arraybackslash}X@{}}
% \toprule
% Maneuver & Path characteristics & Lateral--yaw excitation\\
% \midrule
% Lane change
% & Single lateral transition
% & Transient lateral tracking\\
% Double lane change
% & Consecutive alternating transitions
% & Steering reversal and yaw response\\
% Slalom
% & Repeated alternating turns
% & Lateral acceleration and yaw reversals\\
% Circular path
% & Approximately constant curvature
% & Sustained lateral force and yaw response\\
% J-turn
% & Rapid entry into a pronounced turn
% & Transient yaw and sideslip response\\
% U-turn
% & High-curvature turn with large heading change
% & Sustained turning and steering coordination\\
% Variable-curvature cornering
% & Changing path curvature
% & Varying lateral acceleration demand\\
% \bottomrule
% \end{tabularx}
% \end{table}
Only successful expert rollouts were retained, requiring maneuver completion, valid MPC solutions, and bounded vehicle responses. The resulting dataset comprised $181$ trajectories, divided into training and validation subsets at the trajectory level with a $75{:}25$ ratio to prevent temporal data leakage.

The policy input comprised $22$ vehicle-state, tracking-error, actuator, and tire-related features, together with $4H$ path-preview features, where $H=10$. The network predicted $H_c=10$ front--rear steering pairs supervised by the corresponding expert MPC sequences. Training employed AdamW with a batch size of $128$, followed by $160$ imitation-learning epochs and $340$ safety-guided fine-tuning epochs, as detailed in Table~\ref{tab:training_loss_parameters}.

Generalization tests were conducted on held-out trajectory, speed, and adhesion conditions excluded from policy training. LQR, MPC, and the proposed controller were evaluated on identical scenarios. All three controllers completed the prescribed generalization tests, and the statistics in Table~\ref{tab:generalization_comparison} were calculated without controller-specific success filtering. Separate near-limit and emergency maneuvers were used for handling-response analysis and ablation studies. CarSim provided the nonlinear vehicle response, while the safety-projection layer used the reduced-order model introduced in Section~\ref{sec:idcbf}.

\subsection{Baselines and Evaluation Metrics}
Two model-based controllers were considered as baselines. The LQR controller provided an unconstrained linear feedback reference, and the MPC controller represented the model-based trajectory-tracking benchmark. The controllers were evaluated using identical test scenarios within each generalization category. Maximum and mean absolute lateral displacement errors quantify tracking accuracy, while the maximum and mean magnitudes of vehicle sideslip angle and yaw rate characterize the handling response. Larger sideslip and yaw-rate magnitudes are not themselves proof of unsafe behavior. Safety-related interpretations also require the applicable handling boundaries and tire-force limitations.

\subsection{Generalization to Unseen Operating Conditions}
\label{subsec:generalization}
Three held-out test groups were designed to evaluate generalization across trajectory, velocity, and road-adhesion conditions. Trajectory tests used unseen lane-change geometries at $30$--$60~\mathrm{km/h}$ and $\mu=0.6$--$1.0$. Velocity tests extended the reference speed to $65$--$80~\mathrm{km/h}$, while adhesion tests reduced $\mu$ to $0.3$--$0.5$. All controllers were evaluated under identical conditions.

As reported in Table~\ref{tab:generalization_comparison}, the proposed method reduced the mean lateral tracking error from $0.0358$ to $0.0199~\mathrm{m}$ in trajectory generalization and from $0.0449$ to $0.0231~\mathrm{m}$ in adhesion generalization, compared with MPC. Under velocity generalization, it achieved a lower maximum error of $0.1783~\mathrm{m}$, although its mean error of $0.0443~\mathrm{m}$ was slightly higher than the LQR result. The proposed controller generally exhibited larger sideslip and yaw-rate responses than MPC, motivating further evaluation of near-limit handling behavior.

\subsection{Near-Limit Handling Assessment}
A low-adhesion J-turn was conducted to evaluate tracking and handling responses near the stability limits. As shown in Fig.~\ref{fig:flag104_test}, LQR exhibited substantial trajectory deviation and unstable lateral--yaw responses. MPC mitigated the instability but showed slower tracking convergence. The proposed controller completed the maneuver with faster convergence while exhibiting transient excursions beyond the prescribed handling boundary. These results demonstrate improved tracking--stability coordination in this near-limit scenario without implying strict constraint satisfaction.

\subsection{Ablation Study of Safety Projection and Learning}
High-speed single- and double-lane-change maneuvers were conducted at $80$--$90~\mathrm{km/h}$ and $\mu=0.6$--$1.0$ to evaluate the contributions of safety projection and policy training.

As shown in Table~\ref{tab:ablation_study}, the fixed non-differentiable safety filter failed to complete the prescribed maneuvers, with maximum lateral errors exceeding $3~\mathrm{m}$. The differentiable safety layer without joint training completed the maneuvers but produced maximum and mean lateral errors of $1.8739$ and $0.4981~\mathrm{m}$. The nominal policy achieved corresponding errors of $0.4562$ and $0.1150~\mathrm{m}$. The complete framework reduced these errors to $0.3422$ and $0.1011~\mathrm{m}$, demonstrating the benefit of coordinating nominal policy learning with safety-related supervision and projection under the evaluated emergency conditions.

\section{Discussion}
\label{sec:discussion}

The proposed differentiable chassis safety layer provides a modular interface between task-oriented control and explicit physical constraints. Although demonstrated through trajectory tracking, the architecture can be extended to other vehicle motion-control objectives with appropriate task formulations. The generalization and near-limit experiments demonstrate improved tracking performance under the evaluated conditions, while the ablation results highlight the benefits of coordinating policy learning with chassis safety projection.

The dynamics-unrolled formulation captures the temporal coupling between projected control actions and subsequent safety constraints. However, the effectiveness of projection-dependent gradients depends on the training configuration. Furthermore, the simplified vehicle model, relaxed safety constraints, and assumed state availability preclude strict closed-loop safety guarantees. Future work will investigate uncertainty-aware safety projection, real-time implementation, and experimental validation.

\section{Conclusion}
\label{sec:conclusion}

This article proposed a differentiable chassis safety layer for intelligent vehicle motion control. Actuator limits, individual-tire friction constraints, and vehicle handling envelopes are embedded into sequential QPs coupled through predictive vehicle dynamics. The resulting layer enables modular integration with task-oriented controllers and provides projection sensitivities for safety-guided learning. A two-stage training strategy combines expert imitation with risk supervision and conditional intervention guidance. Closed-loop CarSim experiments demonstrate improved task execution under the evaluated generalization and emergency conditions, supporting the effectiveness of the proposed task--safety coordination framework.

\section*{Declarations}
On behalf of all authors, the corresponding author states that there is no conflict of interest.

\section*{Acknowledgments}
This work was supported in part by the National Key R\&D Plan of China (Grant No. 2024YFB2505200), and in part by the National Science Foundation of China under Grants 52472452 and 52325212.

\bibliographystyle{unsrt}
\bibliography{references}

\section*{Author Biographies}
\noindent\begin{minipage}[c]{0.15\linewidth}
\centering
\includegraphics[width=0.94\linewidth,height=1.0in,keepaspectratio]{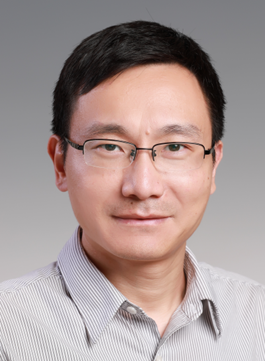}
\end{minipage}\hfill
\begin{minipage}[c]{0.82\linewidth}
\textbf{Lu Xiong} received the Ph.D. degree in vehicle engineering from Tongji University, Shanghai, China, in 2005. From November 2008 to November 2009, he was a Postdoctoral Fellow with the Institute of Automobile Engineering and Vehicle Engines, University of Stuttgart, Germany, with Dr. Jochen Wiedemann. He is now a Professor at Tongji University. He is also the Dean of the College of Automotive and Energy Engineering, the Director of the Clean Energy Automotive Engineering Center, and the Executive Director of the Institute of Intelligent Vehicles, Tongji University. His research interests include perception, decision, and planning; dynamic control; state estimation; and testing and evaluation of autonomous vehicles.
\end{minipage}
\par\medskip
\noindent\begin{minipage}[c]{0.15\linewidth}
\centering
\includegraphics[width=0.94\linewidth,height=1.0in,keepaspectratio]{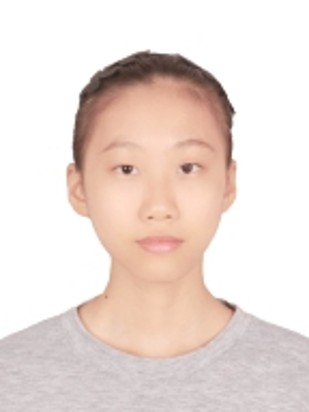}
\end{minipage}\hfill
\begin{minipage}[c]{0.82\linewidth}
\textbf{Xuge Hu} received the B.S. degree in the School of Transportation, Jilin University, China, in 2024. She is currently pursuing the Ph.D. degree in the College of Automotive and Energy Engineering, Tongji University, China. Her research interests include intelligent vehicle control and autonomous driving.
\end{minipage}
\par\medskip
\noindent\begin{minipage}[c]{0.15\linewidth}
\centering
\includegraphics[width=0.94\linewidth,height=1.0in,keepaspectratio]{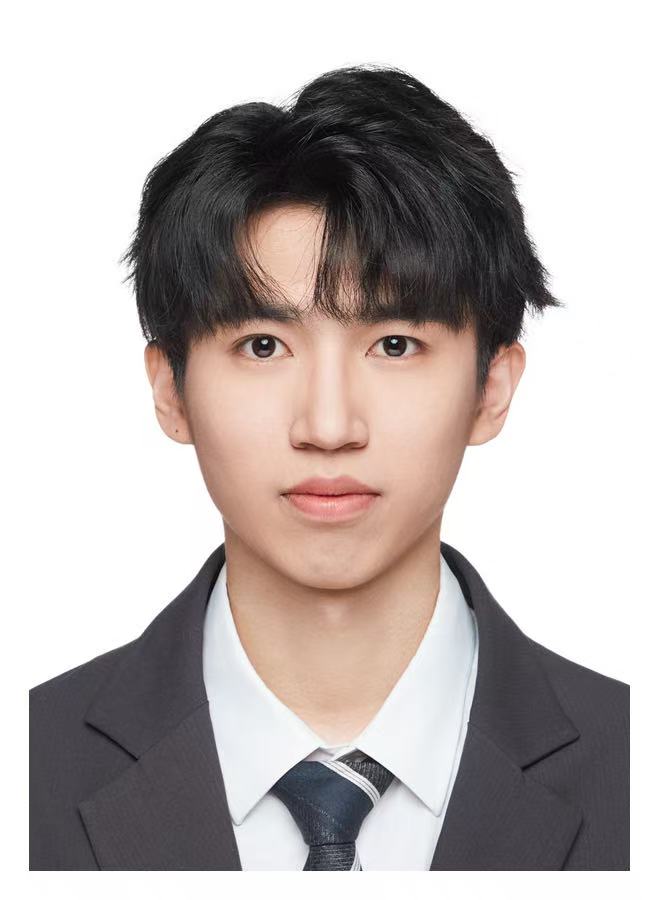}
\end{minipage}\hfill
\begin{minipage}[c]{0.82\linewidth}
\textbf{Yufei Song} received the B.S. and M.S. degrees from the College of Automotive and Energy Engineering, Tongji University, Shanghai, China, in 2023 and 2026, respectively. His research interests include autonomous driving and motion planning.
\end{minipage}
\par\medskip
\noindent\begin{minipage}[c]{0.15\linewidth}
\centering
\includegraphics[width=0.94\linewidth,height=1.0in,keepaspectratio]{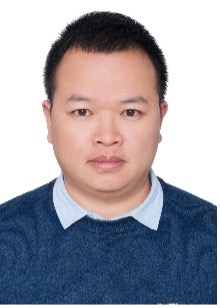}
\end{minipage}\hfill
\begin{minipage}[c]{0.82\linewidth}
\textbf{Weidong Liu} received the M.E. degree from China Agriculture University, Beijing, China, in 2012. He is now a Doctoral student studying in the School of Automotive Studies, Tongji University, and working at Jiangling Motors Co., Ltd. His research interests include decision, planning, and dynamic control of autonomous vehicles.
\end{minipage}
\par\medskip
\noindent\begin{minipage}[c]{0.15\linewidth}
\centering
\includegraphics[width=0.94\linewidth,height=1.0in,keepaspectratio]{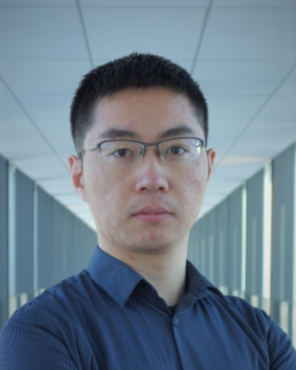}
\end{minipage}\hfill
\begin{minipage}[c]{0.82\linewidth}
\textbf{Chen Tang} received his Ph.D. degree in mechanical and mechatronics engineering from the University of Waterloo, Ontario, Canada, in 2018. From 2019 to 2021, he worked at the University of Waterloo, Waterloo, Canada, as a postdoctoral researcher. He is currently an Associate Professor with the College of Automotive and Energy Engineering, Tongji University, Shanghai. His research interests include advanced chassis systems, vehicular active safety systems, and intelligent transportation systems.
\end{minipage}
\par\medskip
\end{document}